\documentclass[a4paper]{article}

\usepackage{INTERSPEECH2019}

\title{Scalable Multi Corpora Neural Language Models for ASR}
\name{Anirudh Raju, Denis Filimonov, Gautam Tiwari, Guitang Lan, Ariya Rastrow}
\address{
  Amazon Alexa}
\email{\{ranirudh,denf,tgautam,guital,arastrow\}@amazon.com}

\begin{document}
\maketitle
\begin{abstract}

Neural language models (NLM) have been shown to outperform conventional n-gram language models by a substantial margin in Automatic Speech Recognition (ASR) and other tasks. There are, however, a number of challenges that need to be addressed for an NLM to be used in a practical large-scale ASR system. 
In this paper, we present solutions to some of the challenges, including training NLM from heterogenous corpora, limiting latency impact and handling personalized bias in the second-pass rescorer.
Overall, we show that we can achieve a 6.2\% relative WER reduction using neural LM in a second-pass n-best rescoring framework with a minimal increase in latency.
\end{abstract}


\section{Introduction}
Language Models (LM) are a key component in building Automatic Speech Recognition (ASR) systems. The most common approach to building LMs for ASR systems is to learn back-off n-gram models on large text corpora. Recurrent Neural Language Models (NLM) have been shown to consistently outperform traditional n-gram language models from language modeling benchmarks \cite{bengio2003a, schwenk2007continuous, mikolov2010recurrent, jozefowicz2016exploring}.

It is challenging to incorporate NLMs directly into ASR decoding partly because, unlike n-gram models, they model unlimited context history, resulting in exponential explosion in the decoder search space,
and partly because in a large vocabulary ASR system the number of acoustically plausible hypotheses can be very large. There has been prior work in the area to address this issue \cite{hori2014real} but it is still computationally very expensive. Alternatively, a more common approach is to run the ASR decoding in two passes, where the first-pass ASR system is decoded with an n-gram language model to generate a pruned search space in the form of a word lattice or an n-best hypothesis list. In the second-pass, this pruned word lattice is rescored with a stronger neural language model. One of the drawbacks of using a two-pass strategy for a real-time streaming ASR system is that the computation in the second-pass of rescoring is performed after the completion of the streaming first-pass decoding. Hence, this additional computation in the second-pass manifests itself in the form of pure latency for the ASR system. Moreover, an additional drawback is that the stronger NLM is not used during first-pass decoding, which potentially results in losing good hypotheses due to beam search.

Our main contribution in this paper is to call out and address several challenges in bringing NLMs into a latency sensitive ASR system.  In particular, the challenges that we address are 
\begin{itemize}
\item Training the NLM on multiple heterogenous corpora, effectively a domain adaptation problem 
\item Incorporating the NLM into the ASR system, while limiting the latency impact. This includes strategies both to reduce the second-pass NLM computation and to get benefit from the NLM in the first-pass of decoding

\item Pesonalizing NLM by passing biases for classes such as contact names from the first-pass model through the NLM
\end{itemize}

The rest of the paper is organized as follows. Section \ref{section:challenges} describes our proposed method to tackle each of challenges that we address in this work. In particular, in Section \ref{section:domain-adaptation}, we describe our solution to the domain adaptation problem, in Section \ref{section:fast-inference} we describe our solution for fast inference including the usage of self-normalized models and quantization. Section \ref{section:generate} describes the usage of synthetic data generated from the NLM, to improve the first-pass model and hence the n-best hypothesis list used in second-pass rescoring.

 In Section \ref{section:special-token-handling}, we talk about how we handle neural rescoring when the first-pass model is a class based n-gram model \cite{brown1992class} with personalized biases. In Section \ref{experimental-setup}, we describe our experimental setup and in Section \ref{section:results-and-discussion}, we dive into our results. We conclude in Section \ref{section:conclusions} and outline directions for future work.

\section{Methods and Challenges Addressed}
	\label{section:challenges}

\subsection{Domain adaptation}
	\label{section:domain-adaptation}

In a practical ASR system, the LM is often trained on multiple heterogenous corpora, comprising a mix of written corpora and manually transcribed spoken text corpora from various domains.
These corpora may differ in terms of their vocabulary, content, style, argots, etc \cite{bellegarda2004statistical}.
We require a solution to train a neural LM on such heterogenous training data. We present two approaches below in Sections \ref{section:data-mixing} and \ref{section:transfer-learning} to deal with this.

\subsubsection{Data mixing}
	\label{section:data-mixing}
Training an n-gram language model on a variety of diverse corpora is straightforward. A typical strategy is to train separate n-gram models on each corpus, and combine them through linear interpolation with weights optimized to minimize perplexity on an in-domain development set \cite{gandhe2018scalable, raju2018contextual}. N-gram models have the benefit that the final linearly interpolated model, is also represented as an n-gram model. This allows for easy integration of the interpolated model into the ASR system. NLMs however, require a different approach to learning from heterogenous corpora, as a linear interpolation of NLMs, results in an ensemble model with much higher computation.

In this paper, we propose a novel solution to this problem which is simple yet effective. Parameters of neural networks are typically estimated using a variant of stochastic gradient descent, and this method relies on each minibatch being an Independent and Identically Distributed (iid) sample of the distribution we are trying to learn. Thus, we construct minibatches stochastically, by drawing samples from each corpus with probability according to its relevance weight. This has an advantage over other alternatives such as scaled loss function because it allows the combination of corpora with arbitrarily different sizes and relevance weights, and in a practical system, both can vary by 2-3 orders of magnitude.
For relevance weights, we construct n-gram models from each data source and optimize their linear interpolation weights on a development set.
While these weights are not necessary optimal we found that they work well in practice.
In the future, we plan to investigate methods for learning the relevance weights as part of NLM training procedure.

\subsubsection{Transfer learning through fine tuning}
	\label{section:transfer-learning}

One of the often used approaches to deal with the domain adaptation problem is to use fine-tuning, i.e., to train a neural network on a large out-of-domain dataset and subsequently fine-tune the parameters of the model on an in-domain dataset. Some of the parameters can optionally be fixed during the adaptation, typically those corresponding to the lower layers of the model which learn more generic transformations that are not domain specific. This has been successfully applied in computer vision \cite{krizhevsky2012imagenet,zeiler2014visualizing} and NLP \cite{mou2016transferable}.
The downside to this approach is that it does not leverage the fact that each individual out-of-domain corpus has varying relevance to the target domain. Moreover, the model also faces the challenge of catastrophic forgetting \cite{mccloskey1989catastrophic, ratcliff1990connectionist}, where the model loses past knowledge of the pre-trained weights. In order to get benefits from this method and alleviate some of its drawbacks, this approach can be combined with the data mixing strategy described in Section \ref{section:data-mixing}. The model is first pre-trained on the out-of-domain data, and the data mixing strategy is used during the fine tuning stage.

\subsection{Fast inference solutions}
	\label{section:fast-inference}

\subsubsection{Self-normalized models}
In NLMs, the probability of word $w_{i}$ given it's word history \textbf{h} is given by Eqn. \ref{eqn:self-normalization} below:

 \begin{equation}
   p(w_{i}|\textbf{h}) = \frac{exp(z_{i})}{\sum_{j=1}^{|V|} exp(z_{j})} = \frac{exp(z_{i})}{Z(\textbf{h})}
   \label{eqn:self-normalization}
 \end{equation}

\noindent
where $z_{i}$ is the unnormalized logit corresponding to word $w_{i}$ which is computed as an inner product,
$z_{i}=exp(\textbf{c}^{T}\textbf{e}_{w_{i}} + b_{i})$ where $\textbf{c}^{T}$ is the hidden output context vector and $\textbf{e}_{w_{i}}$, $b_{i}$ are the output word embedding vector and bias value for word $w_{i}$. The normalization term $Z(\textbf{h}) = \sum_{j=1}^{|V|} exp(z_{j})$ is known as the partition function, and it involves a summation over all words in the vocabulary. In large vocabulary NLMs, most of the computation cost is incurred to compute the partition function that produces a proper distribution over the vocabulary as this cost is proportional to the vocabulary size. 

There has been a lot of prior work on reducing this computation cost, during both training time and inference time \cite{mikolov2011extensions,morin2005hierarchical,chen2016efficient,shi2014variance} either by approximating the partition function or by modifying the loss function to encourage the model to learn to produce approximately normalized scores (self-normalization).

The self-normalization approach allows us to compute only the scores for the query words thus eliminating the dependency on the vocabulary size. One of the approaches that can be used to train self-normalized models is to add a regularization term during training which encourages the normalization term of the softmax to be close to one \cite{devlin2014fast,chen2016strategies,andreas2015when}. Alternatively, Noise Contrastive Estimation (NCE) based training results in neural networks with inherent self-normalization properties \cite{mnih2012a,vaswani2013decoding,chen2015recurrent,zoph2016simple}. The self-normalization properties of these two broad strategies are emperically compared in \cite{goldberger2018self}.  While both strategies perform well computation-wise for inference, the NCE method has the benefit that it is faster during training time as well. NCE based training does not require computation of the full softmax at training time resulting in significant training speed-ups, which is independent of the output vocabulary size.  In this work, we use NCE to train the Neural LMs since it has two very desirable properties of speeding up the computation during both training and inference.

\subsubsection{Post-training Quantization}
\label{section:quantization}
Quantization of the weights and activations of trained models to 16-bit fixed-point representation is performed to reduce computational cost during inference time. We perform a per-column quantization of the weight matrices, where different shifts and scales are used for each column. We found that this type of quantization performs better that using a global shift and scale for the entire matrix. This method has similarities to the per-channel quantization for convolutional networks in past work \cite{krishnamoorthi2018quantizing}, which uses a different scale and shift for each convolutional kernel. While other work has explored quantization-aware training to squeeze out lower bit representations without accuracy loss, we leave this to future work.

\subsection{Generating synthetic data for first-pass LM}
\label{section:generate}

There is a major drawback of using an NLM strictly in the second-pass to rescore lattices or n-best lists. A weaker n-gram LM is used in the first-pass and some hypotheses may be pruned, which makes them unrecoverable in the second-pass rescoring. Prior work \cite{hori2014real} has attempted to tackle this problem by incorporating scores from the NLM into the first-pass beam search, however this is computationally expensive.

In our system, we take the approach proposed in \cite{deoras2011variational}, namely we construct an n-gram approximation of NLM by sampling a large text corpus from NLM and estimating an n-gram model from that corpus.
Unlike \cite{deoras2011variational}, however, we use a \emph{subword} NLM to generate synthetic data so that the generated corpus will not be limited to the vocabulary of the current version of the ASR system. Sentences containing out of vocabulary words are discarded.
This way, as the vocabulary changes from version to version, we can re-use the same synthetic data.

\subsection{Handling personalized first-pass LM}
\label{section:special-token-handling}

The first-pass LM may have classes \cite{brown1992class} with personalized biases, for example contact names \cite{aleksic2015improved}. 
An NLM trained on general data, however, would not have good estimates for such highly personalized words or phrases.
In such cases, we trust the scores of the personalized first-pass LM more than the scores of a general NLM and we do the following:
Surround class content with tags \verb|<class>| and \verb|</class>|, and the words between the tags will retain their first-pass LM scores but they are still passed through the NLM in order to update its state so that the words after the closing tag will be estimated using the correct history.

\section{Experimental Setup}

\label{experimental-setup}
In all of the experiments in this paper, we build an ASR system that targets the message dictation task. The ASR system comprises first-pass LM trained on a variety of in- and out-of-domain corpora, including written text data and transcribed speech data. The transcribed speech data is from real user-agent interactions, and is bucketed into two separate categories - (1) message dictation specific data and (2) all other types of user-agent interactions. The transcribed messaging data which comprises of approximately 5 million words of text, is our only in-domain data corpus. 
The written text corpora contain over 50 billion words in total. One corpus is a 150M word long-form voicemail dataset. Although superficially similar, distributionally it is quite different from our task: for example, the average utterance length in this dataset is 67 words while our in-domain transcribed corpus has only 15 words on average.

A Kneser-Ney (KN) \cite{kneser1995improved} smoothed n-gram language model is estimated from each corpus, and the final first-pass LM is a linear interpolation of these component LMs. The interpolation weights are estimated by minimizing the perplexity on target development set, in this case transcribed message dictation utterances. In experiments that use NLM generated data, we estimate a separate KN smoothed LM on the synthesized data. This n-gram LM is used as an additional component in the linear interpolation. 

The NLM used in the second-pass rescoring is trained on the voicemail and message dictation corpus only, leaving out the other larger written text corpora. The reasoning behind this is that, the other corpora have a relatively low weight in the n-gram linear interpolation, and they are not that crucial to our message dictation task. The linear interpolation weights from the KN smoothed n-gram LM are 0.78 and 0.22 respectively for the message transcription and the voicemail task. In experiments which use data mixing to train the NLM, these weights are used as relavance scales for the corresponding corpora.

The NLM architecture is two LSTMP \cite{hochreiter1997long, sak2014long} layers, each comprising 1024 hidden units projected down to a dimension of 512. In addition, there are residual connections \cite{he2016deep} between the layers. The models are quantized to 16-bit fixed-point representation as described in Section \ref{section:quantization}. The NLM is used to rescore 10-best hypotheses generated from first-pass decoding.

From in-domain corpus, we extract the vocabulary of 60k most frequent words. All NLM models use this vocabulary and out of vocabulary tokens are mapped to \verb|<unk>|.
Note that the first-pass ASR system has a larger vocabulary, 160k, plus new words can be introduced via personalized classes.
In rescoring experiments (but not for perplexity computation), we scale the probability of \verb|<unk>| token by a factor of $10^{-5}$, i.e., we assume a uniform distribution over the "missing" vocabulary.

\section{Results and Discussion}
\label{section:results-and-discussion}

\subsubsection{Domain adaptation experiments}
Table \ref{table:data-mixing} shows perplexity results comparing NLMs trained on a single data source against different domain adaptation methods described in Sections \ref{section:data-mixing} and \ref{section:transfer-learning}: mixing multiple corpora, applying transfer learning (fine-tuning), and combining both methods. 
First, we confirm that our voicemail corpus is indeed out-of-domain: perplexity of an NLM model trained on just that data is 116.0, more than double the perplexity of a model trained on in-domain corpus only, 55.8.
Next, we study the impact of the two domain adaptation strategies.
The results from Table \ref{table:data-mixing} show a 12.6\% relative improvement in perplexity using transfer learning compared to a baseline trained on in-domain message dictation data only. The model is trained through fine-tuning i.e. initially learning a model on the out-of-domain voicemail corpus and further fine-tuning on in-domain message dictation data. This is in line with prior work in literature. More interestingly, by training the model directly on data mixed from both messaging and voicemail data with the relevance weights estimated from an interpolated KN smoothed n-gram model, we can obtain a 13.4\% improvement in perplexity compared to the baseline model. These results are very promising, since the models trained with the data mixing approach, provide slightly better perplexity results, and train significantly faster than the transfer learning approach. The disparity in training speeds is because the transfer learning approach requires two rounds of training, with the pre-training round performed on a significantly larger out-of-domain corpus (which is usually the case, since there is far lesser in-domain training data available). In the data-mixing approach, the model converges much quicker, seeing several epochs of the in-domain data and fewer epochs (possibly lesser than one) on the out-of-domain data. This is for the simple reason that the in-domain corpus is much smaller and the sampling weights of the two corpora is typically skewed towards the in-domain corpus (0.78 in our experiments). 

Finally, it is possible to combine the two approaches described above, i.e., first pre-training the model on out-of-domain data and then fine-tuning on a mixture of in-domain and out-of-domain data. This results in a 16.1\% relative improvement in perplexity compared to the baseline, which is better than each of the individual approaches alone. In all future experiments, we use the best NLM obtained from both data-mixing and fine-tuning

\begin{table}[th]
  \caption{Perplexity results for domain adaptation. Voicemail corpus is out-of-domain for the message dictation task. "Mix" refers to the data mixing approach}
  \label{table:data-mixing}
  \centering
    \begin{tabular}{l | l | r }

    \toprule
    \multicolumn{1}{c}{\textbf{Pretrain Corpus}} & \multicolumn{1}{c}{\textbf{Train Corpus}} & \multicolumn{1}{c}{\textbf{PPL}} \\
    \midrule
    -  & Voicemail & 116.0 \\
    -  & Messaging & 55.8 \\
    Voicemail & Messaging & 48.8 \\
    - & Voicemail + Messaging mix & 48.3 \\
    Voicemail & Voicemail + Messaging mix &  46.8 \\
    \bottomrule
  \end{tabular}
  
\end{table}

\subsubsection{Inference speed impact of self-normalized LM}
Table \ref{table:self-normalized-ppl} shows that the perplexity of unnormalized and normalized models are very close, which will allow us to use unnormalized probabilities for the second-pass rescoring saving a bulk of the inference computation time. 
In order to show this, we compare the p50 and p90 percentiles for latency added purely due to the second-pass rescoring. This is shown in Table \ref{table:wer-numbers}, where the rescoring latency of the self-normalized NCE LM is lower than the softmax LM by about 700ms at p50 and 3100ms at p90 percentiles. 

\begin{table}[th]
  \caption{Perplexity results comparing normalized and unnormalized NCE models on a voicemail development set. Unnormalized probabilities do not include the softmax normalization factor}
  \label{table:self-normalized-ppl}
  \centering
    \begin{tabular}{l | l }
    \toprule
    \multicolumn{1}{c|}{\textbf{Model}} & \multicolumn{1}{c}{\textbf{PPL}} \\
    \midrule
    Softmax NLM  & 19.42  \\
    NCE NLM (normalized) & 19.95  \\
    NCE NLM (unnormalized) & 20.44  \\
    \bottomrule
  \end{tabular}
  
\end{table}

\subsubsection{WER Impact from NLM}
Table \ref{table:wer-numbers} shows that we are able to obtain 1.6\% relative WER reduction from using NLM-generated synthetic data. Since this is just an update to the first-pass LM there is no increase in latency. These results are in line with the perplexity improvements seen in Table \ref{table:sampling-nlm} with the inclusion of NLM-generated synthetic data. Note that the NLM used to generate the synthetic data is a subword LM, discarding sentences with out of vocabulary words with respect to the first-pass ASR system. In Table \ref{table:sampling-nlm}, the perplexity number of the NLM reported in the last row is of a softmax word-based LM, included as a fair reference for comparison with the KN smoothed n-gram perplexity numbers. 
Finally, performing a 10-best second-pass rescoring using self-normalized NLM gets us a net relative WER reduction of 6.2\%. Note that the WER reduction from both, the softmax NLM and self-normalized NCE NLM, are very similar and in line with the perplexity numbers of Table \ref{table:self-normalized-ppl}.

\begin{table}[th]
  \caption{Relative Word Error Rate Reduction (WERR) and rescoring latency numbers, showing the effect of including NLM synthetic data and rescoring with softmax vs. unnormalized NCE NLM}
  \label{table:wer-numbers}
  \centering
  \footnotesize
	\begin{tabular}{l | c | r | r | r}   
    \toprule    
    \multicolumn{2}{c|}{\textbf{LM}} & \multicolumn{1}{c|}{\textbf{WERR}} &\multicolumn{2}{c}{\textbf{Rescoring latency}} \\
    \textbf{First-pass} & \textbf{Second-pass}  & & P50 & P90 \\
        \midrule
    Baseline & - &  - & - & - \\ 
    ~+Syn data & - & 1.6\% & -  & - \\    
    ~+Syn data & Softmax NLM & 6.3\% & 767ms & 3396ms\\
    ~+Syn data & NCE NLM & 6.2\% & 65ms & 285ms\\ 
    \bottomrule
  \end{tabular}
  
\end{table}

\begin{table}[th]
  \caption{Perplexity results comparing n-gram LMs with and without NLM generated synthetic data on a message dictation test set. "Msg" refers to transcribed message dictation data, "Synthetic Msg" refers to data generated from NLM and "Others" refers to all other available corpora}
  \label{table:sampling-nlm}
  \centering
  \footnotesize
	\begin{tabular}{l | l |c}   
    \toprule
    \multicolumn{1}{c|}{\textbf{LM}} & \multicolumn{1}{c|}{\textbf{Train data}} &\multicolumn{1}{c}{\textbf{PPL}} \\
    \midrule
    KN-4g & Msg & 63.71 \\
    KN-4g-Interp & Msg + Others & 60.81  \\    
    \midrule    
    KN-4g-Syn & Synthetic Msg & 58.34 \\
    KN-4g-Interp-Syn & Msg + Synthetic Msg + Others & 58.11 \\
    \midrule    
    NLM & Msg & 46.85 \\
    \bottomrule
  \end{tabular}
  
\end{table}

\subsection{Impact from personalized bias from first-pass LM}
\label{section:results-special-tok}
Recognition of contact names is important for a message dictation application. The ASR system in this paper is specifically focused on message dictation payload, where the recognition of contact names, within the message are important from a user experience perspective. For example, a fairly common message such as "hey john how was your day" requires accurate recognition of the name "john", which is challenging for a rare or out of vocabulary name. This benefits from the usage of a class-based LM, with a single class for contact names, since it allows us to use personalized contact names list for biasing the model towards user specific information. 
This is measured through the Entity WER metric. To measure this, we tag each word in our test data using an in-house Named Entity Recognition (NER) tagger. The Entity WER is defined as $(num\_substitutions+num\_deletions)/num\_reference\_words$. The hypothesis and reference are aligned in order to calculate the number of substitutions and deletions corresponding to the tagged reference words. Note that we do not include insertions due to difficulty in attributing whether an insertion error was caused by the entity or the other surrounding words. The results in Table \ref{table:special-tokens} showing the Entity WERR for Person names, demonstrate that by appropriately handling the contacts class in the NLM through class tags, we are able do slightly better than a naive approach of rescoring these with the NLM. These class tags enabled us to induce personalized bias in the rescorer by retaining the first-pass scores for the contact names, ignoring the score from the NLM but using the word input to update the LSTM state information. This method was previously described in detail in Section \ref{section:special-token-handling}. Overall, this enabled accuracy improvements for contact name recognition, which is important to user experience.

\begin{table}[th]
  \caption{Relative Entity Word Error Rate Reduction (WERR) of contact names, with and without personal bias in rescorer}
  \label{table:special-tokens}
  \centering

	\begin{tabular}{l | l | r |}   
    \toprule    
    \multicolumn{2}{c|}{\textbf{LM}} & \multicolumn{1}{c|}{\textbf{Entity}} \\
    \textbf{First-pass} & \textbf{Second-pass}  &   \textbf{WERR(\%)} \\
        \midrule
    KN-4g-Interp-Syn  & - &  - \\
    KN-4g-Interp-Syn & NLM & 9.18\% \\
    KN-4g-Interp-Syn & NLM + bias & 9.56\% \\
    \bottomrule
  \end{tabular}
  
\end{table}

\section{Conclusions and Future Work}
\label{section:conclusions}
In this work, we addressed several challenges for an NLM to be used in a practical large-scale ASR system. In particular, training an NLM from multiple heterogenous corpora using a novel data mixing strategy, along with transfer learning based on fine-tuning that provided 16.1\% relative improvement in perplexity compared to a baseline trained on in-domain data only. Subsequently, we presented work to limit latency impact of the models. The usage of self-normalized LM helped us to reduce the added latency by 700ms and 3100ms at the 50th and 90th percentiles, compared to using softmax based LMs.  We were able to obtain a 1.6\% relative WERR by generating synthetic data from the NLM and incorporating that into an n-gram model used in the first-pass beam search decoding. Overall, this provided a net WERR of 6.2\% relative along with 10-best rescoring. Finally, we showed that we were able to get accuracy improvements for contact names, using personalized list information, by using classes in the first-pass LM and appropriately handling them in the NLM rescoring through class tags. In the future, we plan to evaluate the data mixing strategy in handling more than two corpora as well as investigating methods for optimizing data mixing weights as part of NLM training procedure.

\bibliographystyle{IEEEtran}

\bibliography{mybib}

\end{document}